# A Systematic Evaluation of Imbalance Handling Methods in Biomedical Binary Classification

Jiandong Chen, MS[1], Lingjie Su, MS[2], Le Peng, PHD[2], Yash Travadi, PHD[3], Rui Zhang, PHD[4], Ju Sun, PHD[2]

1 Institute for Health Informatics, University of Minnesota, Minneapolis, MN; 2 Department of Computer Science and Engineering, University of Minnesota, Minneapolis, MN; 3 School of Statistics, University of Minnesota, Minneapolis, MN; 4 Division of Computational Health Sciences, Department of Surgery, University of Minnesota, Minneapolis, MN

Corresponding author: Ju Sun, Ph.D., Department of Computer Science and Engineering, University of Minnesota, 200 Union St SE, Minneapolis, MN 55455; jusun@umn.edu

## Abstract

**Objective:** The primary goal of this study was to systematically examine the impact of commonly used imbalance handling methods (IHMs) on predictive performance in biomedical binary classification, considering the interplay between model complexity and diverse data modalities.

**Material and Methods:** We evaluated five representative IHMs—random undersampling (RUS), random oversampling (ROS), SMOTE, re-weighting (RW), and direct F1-score optimization (DMO)—against a raw training (RAW) baseline. The evaluation encompassed three public biomedical datasets: MIMIC-III (tabular), ADE-Corpus-V2 (text), and MURA (image), spanning three common biomedical data modalities. To assess varying model complexity, we employed a range of architectures, from classical logistic regression and random forest to deep neural networks, including multilayer perceptron (MLP), BiLSTM, BERT, DenseNet, and DINOv2.

**Results:** For simpler models such as logistic regression on tabular data, IHMs yielded no significant advantage over the RAW baseline, aligning with prior findings.[1] However, clear benefits were observed for more complex models and unstructured data: (a) ROS and RW consistently enhanced the performance of powerful models; (b) Direct F1-score optimization demonstrated utility primarily for unstructured text and image data; and (c) RUS and SMOTE consistently degraded performance and are hence not recommended.

**Conclusion:** The effectiveness of IHMs depends on both model complexity and data modality. Performance gains are most pronounced when leveraging appropriate IHMs, such as ROS, RW, and DMO on high-complexity models.

# Introduction

Binary classification models have become indispensable in biomedical research, enabling critical advances in disease diagnosis, risk assessment, and clinical decision support.[1–3] While these tasks typically involve distinguishing between two distinct clinical outcomes, standard learning algorithms often assume a relatively balanced label distribution.[4,5] In practice, however, biomedical datasets are frequently characterized by significant class imbalance, driven by intrinsic population imbalances and the substantial costs associated with acquiring labels for rare conditions.[6,7] In these imbalanced scenarios, conventional empirical risk minimization used for training binary classification models tends to exhibit poor predictive performance on the underrepresented minority class.[8,9] To mitigate this, various imbalance handling methods (IHMs) have been proposed, which can be categorized into three primary families: (a) Data resampling, encompassing random undersampling (RUS), random oversampling (ROS), and the synthetic minority oversampling technique (SMOTE); (b) Cost-sensitive learning, most notably re-weighting (RW) strategies that assign differential importance to samples based on class frequency;[10–12] and (c) Direct metric optimization, which seeks to improve performance by directly maximizing imbalance-sensitive criteria such as the F1-score and precision-recall tradeoff.

Despite their widespread adoption, the effectiveness of IHMs remains a subject of ongoing debate, with prior research yielding divergent results. Several investigations have demonstrated clear advantages: a survey by Johnson and Khoshgoftaar found that ROS consistently outperformed RUS.[13] Similarly, Khushi et al. evaluated 23 IHMs using random forest, logistic regression, and linear SVC on imbalanced lung cancer datasets, revealing that oversampling techniques—specifically ROS—yielded superior predictive performance.[14] Zhao et al. proposed an IHM framework for healthcare data that integrated multiple rebalancing strategies, finding that combining logistic regression with SMOTE yielded optimal results.[6] Furthermore, Travadi et al. and Peng et al. introduced direct metric optimization methods for imbalanced classification, demonstrating that these approaches consistently outperform traditional re-weighting strategies.[15,16] Conversely, other studies have highlighted potential drawbacks associated with IHMs. Van den Goorbergh et al. examined the impact of RUS, ROS, and SMOTE on logistic regression using both simulated and real-world biomedical data.[17] Notably, they observed that IHMs often exerted a detrimental effect, distorting model calibration and leading to overconfident predictions that reduced clinical utility without improving overall discrimination.[18] However, that investigation was limited to logistic regression and tabular data, potentially overlooking the impact on more powerful models and unstructured data modalities. Additionally, their analysis relied on a restricted set of metrics and a fixed classification threshold of 0.5, which may not be optimal for imbalanced settings.

The primary goal of this study was to systematically examine common imbalance handling methods (IHMs) (RUS, ROS, SMOTE, RW, DMO) and compare them against the raw training (RAW) baseline in binary classification tasks. While prior studies have often been limited to a single modality, we extended our evaluation to encompass three distinct data modalities—tabular, text, and image data—increasingly prevalent in biomedical research. Furthermore, we employed a diverse set of models spanning varying levels of complexity, from classical ones such as logistic regression and random forest to more powerful ones based on deep neural networks. Specifically, we evaluated logistic regression, random forest, and MLP for tabular data; BiLSTM and BERT for text data; and DenseNet and DINOv2 for image data. Our findings indicate that the effectiveness of IHMs depends on both model complexity and data modality, with ROS and RW emerging as the most consistently beneficial approaches. This research offers critical insights into imbalanced biomedical binary classification, helping practitioners select appropriate IHMs across diverse domains.

# Method

## Data Sources

**Medical Tabular Data**: The dataset utilized in this evaluation was derived from the widely adopted MIMIC-III database,[19] an extensive, de-identified repository of health-related data from critical care admissions.[20–24] We leveraged MIMIC-Extract, a standardized preprocessing pipeline, to generate the final analytical cohort.[25] The primary prediction target was ICU mortality. For patients with longitudinal records, features were aggregated at the patient level using mean values across observations. The feature set encompassed: (a) demographic indicators, including age, gender, ethnicity, and maximum length of stay; (b) mean physiological vital signs and laboratory results; and (c) clinical interventions. To handle the missing data inherent in MIMIC-III, features with a missing rate exceeding 25% were excluded, leaving 82 continuous features. Two imputation strategies were compared: (a) MeanImputer, replacing missing values with feature-wise means, and (b) IterativeImputer, employing multivariate imputation via the *scikit-learn* Python package.[26] The final dataset was characterized by significant class imbalance, with a positive-to-negative ratio of 1:14.2. We evaluated four models spanning varying degrees of complexity: logistic regression, ridge logistic regression, random forest, and a multilayer perceptron (MLP).

**Medical Text Data**: The dataset analyzed for this study was the ADE-Corpus-V2, a publicly available dataset derived from *Medline* case reports. We utilized the sentence-level adverse drug effect (ADE) classification subset, which provides binary labels indicating whether a sentence describes an ADE. The corpus exhibits notable class imbalance, with a positive-to-negative sample ratio of approximately 1:9.8. To capture varying model complexity, we evaluated two architectures: a bidirectional LSTM (BiLSTM) and a pretrained BERT model, representing classical recurrent and state-of-the-art transformer-based models, respectively.

**Medical Image Data**: The dataset utilized in this analysis was MURA (Musculoskeletal Radiographs),[27] a comprehensive public dataset comprising bone X-rays across seven standard upper-extremity study types: elbow, finger, forearm, hand, humerus, shoulder, and wrist. The primary objective was to classify X-ray studies as normal or abnormal. As the original corpus did not exhibit extreme imbalance, we simulated class imbalance by randomly downsampling the minority class to achieve a positive-to-negative sample ratio of approximately 1:5. To evaluate the impact of model complexity, we employed two deep learning models: DenseNet, a convolutional neural network (CNN), and DINOv2, a state-of-the-art vision transformer.[28,29]

The detailed information for all three datasets used in the experiments is shown in Table 1.

**Table 1**. Dataset composition and class distribution across modalities.

| Modality | Dataset | Count | Imbalanced ratio |
|---|---|---|---|
| Tabular Data (MIMIC ICU Mortality) | Training | 25767 : 1809 | 14.2 : 1 |
| | Validation | 3222 : 226 | 14.2 : 1 |
| | Testing | 3222 : 226 | 14.2 : 1 |
| Text Data (ADE Corpus V2) | Training | 13356 : 1364 | 9.8 : 1 |
| | Validation | 1670 : 170 | 9.8 : 1 |
| | Testing | 1669 : 171 | 9.8 : 1 |
| Image Data (MURA) | Training | 19741 : 3948 | 5 : 1 |
| | Validation | 2194 : 439 | 5 : 1 |
| | Testing | 1667 : 333 | 5 : 1 |

## Imbalance Handling Methods

We evaluated five representative imbalance handling methods (IHMs)—random undersampling (RUS), random oversampling (ROS), SMOTE, re-weighting (RW), and direct F1-score optimization (DMO)—against a raw training (RAW) baseline. In binary classification, RUS achieves class balance by randomly discarding samples in the majority class.[18,30] Conversely, ROS augments the minority class by resampling its samples with replacement to match the majority class's size. SMOTE, a synthetic oversampling technique, generates new minority class samples by interpolating between existing instances and their nearest neighbors.[11] RW employs a cost-sensitive learning strategy, typically assigning sample weights inversely proportional to class frequencies to emphasize the minority class during training.[31,32] Unlike these strategies, DMO addresses imbalance by directly maximizing a target performance metric; in this study, we targeted the F1 score, which represents the precision-recall tradeoff, under the DMO framework. To perform gradient-based training, we optimized a differentiable surrogate of the F1-score following Le et al.[16]

To ensure a comprehensive evaluation, RUS, ROS, and RW were systematically applied across all three data modalities. In contrast, DMO was implemented exclusively for neural network–based architectures—including MLP, BiLSTM, BERT, DenseNet, and DINOv2—and was omitted for classical models such as logistic regression and random forest. Furthermore, SMOTE was restricted to the tabular domain, as defining a coherent feature space and generating synthetic instances via interpolation makes it less suitable for high-dimensional, unstructured data.[33,34]

## Evaluation Metrics

Given the imbalanced nature of the datasets, performance was evaluated primarily using F1-score, AUROC (Area Under the Receiver Operating Characteristic Curve), and AUPRC (Area Under the Precision-Recall Curve), [35] as they are more responsive to true classification performance. While AUROC characterizes the tradeoff between true-positive and false-positive rates across varying thresholds, AUPRC provides a more granular depiction of the precision-recall tradeoff, [36] and is considered a more informative criterion for evaluating binary classifiers under significant class imbalance.[37–41] Additional metrics, including accuracy, sensitivity (recall), specificity, and precision, are also reported as secondary metrics for completeness and as occasional tie-breakers.

## Training and Testing

For tabular data, the dataset was partitioned into training (80%), validation (10%), and test (10%) sets. We evaluated four models—logistic regression, ridge logistic regression, random forest, and a five-layer multilayer perceptron (MLP) (Figure 1a). We partitioned the text dataset similarly and evaluated a bidirectional LSTM (BiLSTM) trained from scratch and a pretrained BERT model to assess the impact of model complexity (Figure 1b). For the image dataset, we evaluated the DenseNet-169 and DINOv2 models using both linear probing and full fine-tuning strategies (Figure 1c). Across all data modalities, hyperparameter tuning and model selection were based on validation performance for each IHM, with the respective loss used as the selection criterion. Classification thresholds were optimized on the validation set to maximize the F1-score, rather than relying on a fixed threshold of 0.5. Final performance metrics were reported on the held-out test set, with all experiments repeated across five independent runs to estimate variability.

Detailed model configurations and hyperparameter search spaces are provided in the Supplement.

**Figure 1.** Imbalance handling methods (IHMs) and model architectures used for different data modalities

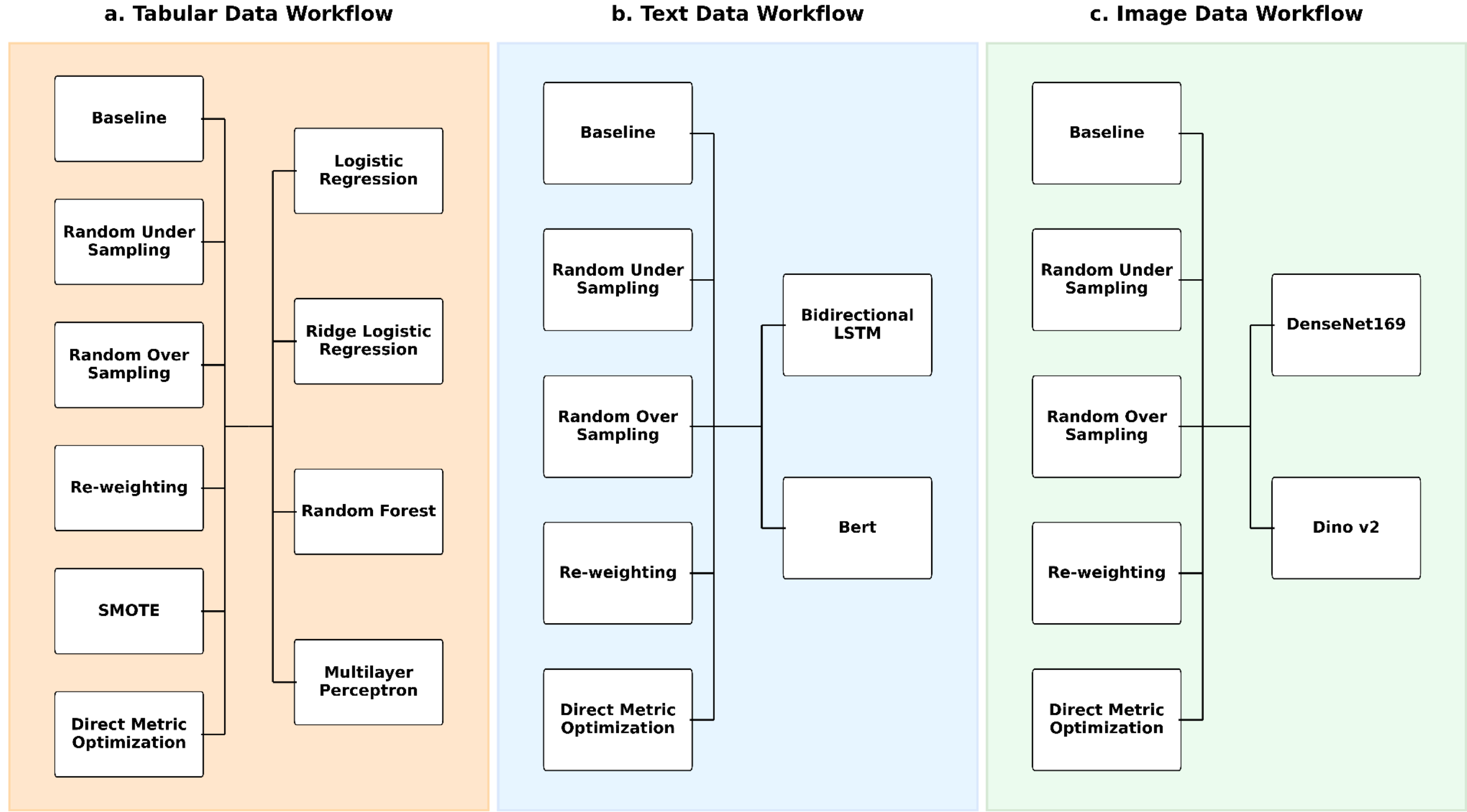

# Results

## Medical Tabular Data: MIMIC-III

The quantitative results for ICU mortality prediction using the MIMIC-III dataset are summarized in Table 2. Performance with mean imputation for missing data consistently surpassed that of IterativeImputer; consequently, our analysis focuses on results based on mean imputation. In classical logistic regression, optimal performance was achieved with the RAW baseline (F1: 0.677; AUPRC: 0.739; AUROC: 0.941), whereas all IHMs yielded inferior metrics. Similar trends were observed for ridge logistic regression, where the RAW baseline (F1, 0.678; AUPRC, 0.740; AUROC, 0.940) outperformed all IHMs. Notably, IHMs resulted in lower F1-scores and AUPRC values, whereas AUROC remained relatively stable across methods. In contrast, for random forest, both RW (0.770) and ROS (0.764) improved the F1-score over the RAW baseline (0.755). However, RUS and SMOTE consistently degraded F1-score, AUROC, and AUPRC. Regarding deep neural networks (MLP), ROS yielded a higher F1-score than RAW (0.725 vs. 0.718), but a lower AUPRC (0.723 vs. 0.747), while RW, RUS, SMOTE, and DMO consistently reduced performance across all primary metrics.

## Medical Text Data: ADE-Corpus-V2

Table 3 presents the quantitative results for the ADE-Corpus-V2 text dataset. For BiLSTM, performance was highest under RAW (F1, 0.567; AUPRC, 0.599; AUROC, 0.897). All IHMs resulted in lower performance, with the largest decrease observed for RUS and DMO. For BERT, DMO achieved the highest F1-score (0.781 vs 0.743 for RAW), followed by ROS (0.780) and RW (0.758). While RW achieved the highest AUPRC (0.810 vs. 0.800 for RAW), ROS and DMO yielded lower AUPRC values than the baseline. RUS consistently degraded all primary metrics.

## Medical Image Data: MURA

Table 4 summarizes the quantitative performance for the MURA medical image dataset. For DenseNet-169, ROS achieved a higher F1-score relative to the RAW baseline (0.755 vs. 0.750), though it yielded a slightly lower AUPRC (0.871 vs. 0.876). Both RW and DMO resulted in lower F1-scores than the RAW baseline, with AUPRC values that were comparable or marginally lower. Notably, RUS consistently degraded F1-score, AUPRC, and AUROC. For DINOv2, DMO attained the highest F1-score (0.733 vs. 0.725 for RAW), albeit with a reduction in AUPRC (0.806 vs. 0.870). RW yielded a modest increase in F1-score (0.727) while maintaining a similar AUPRC (0.868). Conversely, RUS led to a decrease in overall predictive performance.

**Table 2.** Evaluation results for MIMIC ICU mortality prediction (EHR). Values significantly surpassing the benchmark of "RAW" are shown in **red bold**, while values below are shown in **black bold**.

| Class | Model | Imbalance correction methods | Accuracy | Sensitivity (Recall) | Specificity | Precision | F1-score | AUROC | AUPRC |
|---|---|---|---|---|---|---|---|---|---|
| MIMIC ICU Mortality – Missing Imputation (Mean) | Standard Logistic Regression | RAW | 0.962 ± 0.000 | 0.602 ± 0.000 | 0.988 ± 0.000 | 0.773 ± 0.000 | 0.677 ± 0.000 | 0.941 ± 0.000 | 0.739 ± 0.000 |
| | | RUS | 0.932 ± 0.000 | 0.522 ± 0.000 | 0.960 ± 0.000 | 0.480 ± 0.000 | **0.500 ± 0.000** | **0.887 ± 0.000** | **0.511 ± 0.000** |
| | | ROS | 0.949 ± 0.000 | 0.624 ± 0.000 | 0.972 ± 0.000 | 0.610 ± 0.000 | **0.617 ± 0.000** | **0.930 ± 0.000** | **0.664 ± 0.000** |
| | | SMOTE | 0.947 ± 0.000 | 0.588 ± 0.000 | 0.972 ± 0.000 | 0.594 ± 0.000 | **0.591 ± 0.000** | **0.919 ± 0.000** | **0.620 ± 0.000** |
| | | RW | 0.953 ± 0.000 | 0.513 ± 0.000 | 0.984 ± 0.000 | 0.695 ± 0.000 | **0.590 ± 0.000** | **0.924 ± 0.000** | **0.644 ± 0.000** |
| | Ridge Logistic Regression | RAW | 0.963 ± 0.000 | 0.602 ± 0.000 | 0.988 ± 0.000 | 0.777 ± 0.000 | 0.678 ± 0.000 | 0.940 ± 0.000 | 0.740 ± 0.000 |
| | | RUS | 0.948 ± 0.000 | 0.650 ± 0.000 | 0.969 ± 0.000 | 0.595 ± 0.000 | **0.622 ± 0.000** | 0.940 ± 0.000 | **0.692 ± 0.000** |
| | | ROS | 0.956 ± 0.000 | 0.584 ± 0.000 | 0.982 ± 0.000 | 0.698 ± 0.000 | **0.636 ± 0.000** | 0.941 ± 0.000 | **0.707 ± 0.000** |
| | | SMOTE | 0.950 ± 0.000 | 0.704 ± 0.000 | 0.967 ± 0.000 | 0.600 ± 0.000 | **0.648 ± 0.000** | 0.941 ± 0.000 | **0.708 ± 0.000** |
| | | RW | 0.949 ± 0.000 | 0.730 ± 0.000 | 0.964 ± 0.000 | 0.587 ± 0.000 | **0.651 ± 0.000** | 0.940 ± 0.000 | **0.706 ± 0.000** |
| | Random Forest | RAW | 0.965 ± 0.003 | 0.818 ± 0.031 | 0.976 ± 0.005 | 0.703 ± 0.033 | 0.755 ± 0.012 | 0.977 ± 0.001 | 0.822 ± 0.003 |
| | | RUS | 0.966 ± 0.001 | 0.742 ± 0.032 | 0.982 ± 0.003 | 0.741 ± 0.024 | **0.741 ± 0.008** | 0.973 ± 0.001 | **0.794 ± 0.007** |
| | | ROS | 0.967 ± 0.008 | 0.799 ± 0.053 | 0.979 ± 0.012 | 0.740 ± 0.084 | **0.764 ± 0.031** | 0.977 ± 0.001 | 0.821 ± 0.003 |
| | | SMOTE | 0.960 ± 0.003 | 0.565 ± 0.037 | 0.988 ± 0.003 | 0.770 ± 0.043 | **0.650 ± 0.023** | **0.961 ± 0.003** | **0.734 ± 0.004** |
| | | RW | 0.969 ± 0.003 | 0.792 ± 0.041 | 0.981 ± 0.006 | 0.752 ± 0.048 | **0.770 ± 0.006** | 0.977 ± 0.001 | 0.824 ± 0.003 |
| | Deep Neural Network | RAW | 0.965 ± 0.002 | 0.677 ± 0.077 | 0.986 ± 0.004 | 0.770 ± 0.035 | 0.718 ± 0.031 | 0.959 ± 0.002 | 0.747 ± 0.010 |
| | | RUS | 0.958 ± 0.001 | 0.615 ± 0.030 | 0.982 ± 0.002 | 0.708 ± 0.021 | **0.658 ± 0.014** | **0.950 ± 0.003** | **0.675 ± 0.018** |
| | | ROS | 0.963 ± 0.002 | 0.735 ± 0.048 | 0.979 ± 0.004 | 0.717 ± 0.032 | **0.725 ± 0.017** | 0.956 ± 0.003 | **0.723 ± 0.031** |
| | | SMOTE | 0.960 ± 0.003 | 0.695 ± 0.052 | 0.979 ± 0.006 | 0.699 ± 0.048 | **0.695 ± 0.012** | 0.955 ± 0.004 | **0.721 ± 0.013** |
| | | RW | 0.963 ± 0.002 | 0.670 ± 0.038 | 0.984 ± 0.004 | 0.748 ± 0.039 | **0.705 ± 0.005** | 0.960 ± 0.001 | **0.737 ± 0.011** |
| | | DMO | 0.961 ± 0.003 | 0.659 ± 0.027 | 0.982 ± 0.003 | 0.720 ± 0.034 | **0.688 ± 0.023** | **0.871 ± 0.049** | **0.616 ± 0.028** |

**Table 3.** Evaluation results for ADE Corpus V2 (Text data). Values significantly surpassing the benchmark of "RAW" are shown in **red bold**, while values below are shown in **black bold**.

| Model | Imbalance correction methods | Accuracy | Sensitivity (Recall) | Specificity | Precision | F1-score | AUROC | AUPRC |
|---|---|---|---|---|---|---|---|---|
| BiLSTM | RAW | 0.919 ± 0.013 | 0.573 ± 0.065 | 0.954 ± 0.020 | 0.571 ± 0.077 | 0.567 ± 0.026 | 0.897 ± 0.006 | 0.599 ± 0.026 |
| | RUS | 0.898 ± 0.006 | 0.451 ± 0.052 | 0.943 ± 0.006 | 0.448 ± 0.033 | **0.449 ± 0.040** | **0.834 ± 0.014** | **0.414 ± 0.049** |
| | ROS | 0.928 ± 0.004 | 0.505 ± 0.038 | 0.971 ± 0.008 | 0.646 ± 0.054 | 0.565 ± 0.014 | **0.867 ± 0.010** | **0.591 ± 0.025** |
| | RW | 0.915 ± 0.004 | 0.565 ± 0.026 | 0.951 ± 0.007 | 0.540 ± 0.025 | **0.551 ± 0.010** | **0.889 ± 0.006** | **0.550 ± 0.016** |
| | DMO | 0.887 ± 0.039 | 0.481 ± 0.038 | 0.928 ± 0.043 | 0.437 ± 0.140 | **0.450 ± 0.082** | **0.798 ± 0.046** | **0.352 ± 0.107** |
| Bert | RAW | 0.955 ± 0.005 | 0.707 ± 0.066 | 0.980 ± 0.008 | 0.789 ± 0.056 | 0.743 ± 0.030 | 0.970 ± 0.001 | 0.800 ± 0.016 |
| | RUS | 0.946 ± 0.006 | 0.752 ± 0.030 | 0.966 ± 0.010 | 0.694 ± 0.052 | **0.720 ± 0.018** | **0.960 ± 0.007** | **0.727 ± 0.040** |
| | ROS | 0.960 ± 0.006 | 0.768 ± 0.043 | 0.979 ± 0.005 | 0.793 ± 0.041 | **0.780 ± 0.032** | **0.953 ± 0.010** | **0.783 ± 0.058** |
| | RW | 0.953 ± 0.005 | 0.791 ± 0.017 | 0.970 ± 0.005 | 0.728 ± 0.035 | **0.758 ± 0.024** | 0.971 ± 0.003 | **0.810 ± 0.023** |
| | DMO | 0.960 ± 0.004 | 0.774 ± 0.021 | 0.979 ± 0.006 | 0.790 ± 0.044 | **0.781 ± 0.011** | **0.900 ± 0.021** | **0.737 ± 0.065** |

**Table 4.** Evaluation results for MURA (Image data). Values significantly surpassing the benchmark of "RAW" are shown in **red bold**, while values below are shown in **black bold**.

| Model | Imbalance correction methods | Accuracy | Sensitivity (Recall) | Specificity | Precision | F1-score | AUROC | AUPRC |
|---|---|---|---|---|---|---|---|---|
| Densenet169 | RAW | 0.795 ± 0.005 | 0.644 ± 0.019 | 0.933 ± 0.009 | 0.898 ± 0.010 | 0.750 ± 0.010 | 0.866 ± 0.003 | 0.876 ± 0.003 |
| | RUS | 0.773 ± 0.013 | 0.619 ± 0.010 | 0.914 ± 0.032 | 0.870 ± 0.039 | **0.723 ± 0.009** | **0.847 ± 0.017** | **0.855 ± 0.016** |
| | ROS | 0.794 ± 0.005 | 0.661 ± 0.031 | 0.917 ± 0.019 | 0.880 ± 0.021 | **0.755 ± 0.014** | 0.861 ± 0.002 | **0.871 ± 0.003** |
| | RW | 0.788 ± 0.008 | 0.633 ± 0.024 | 0.930 ± 0.010 | 0.893 ± 0.011 | **0.741 ± 0.014** | 0.862 ± 0.007 | **0.870 ± 0.007** |
| | DMO | 0.786 ± 0.008 | 0.616 ± 0.029 | 0.943 ± 0.012 | 0.909 ± 0.013 | **0.734 ± 0.017** | 0.861 ± 0.003 | **0.872 ± 0.004** |
| Dino v2 | RAW | 0.781 ± 0.006 | 0.610 ± 0.022 | 0.938 ± 0.011 | 0.900 ± 0.013 | 0.725 ± 0.012 | 0.859 ± 0.006 | 0.870 ± 0.005 |
| | RUS | 0.770 ± 0.006 | 0.602 ± 0.024 | 0.924 ± 0.018 | 0.880 ± 0.021 | **0.714 ± 0.012** | **0.849 ± 0.005** | **0.858 ± 0.006** |
| | ROS | 0.778 ± 0.004 | 0.604 ± 0.023 | 0.937 ± 0.015 | 0.898 ± 0.018 | 0.722 ± 0.011 | 0.858 ± 0.004 | **0.863 ± 0.004** |
| | RW | 0.781 ± 0.012 | 0.609 ± 0.037 | 0.939 ± 0.016 | 0.902 ± 0.018 | 0.727 ± 0.023 | 0.860 ± 0.008 | 0.868 ± 0.004 |
| | DMO | 0.787 ± 0.003 | 0.610 ± 0.009 | 0.950 ± 0.006 | 0.919 ± 0.009 | **0.733 ± 0.005** | **0.825 ± 0.003** | **0.806 ± 0.004** |

# Discussion

Across the experimental evaluation, random oversampling (ROS) and re-weighting (RW) emerged as the most consistently beneficial approaches for enhancing the F1-score. These performance gains were most pronounced when leveraging more powerful models, including random forests, neural networks, and transformer-based models; conversely, simpler models such as logistic regression yielded no significant advantage from imbalance handling. This observation indicates that the effectiveness of IHMs depends heavily on model complexity.

Direct F1-score optimization (DMO) demonstrated inconsistent utility, with benefits observed primarily for unstructured text and image data. In contrast, random undersampling (RUS) and SMOTE consistently degraded predictive performance across varying model complexities and data modalities and are hence not recommended.

Notably, IHMs frequently led to a reduction in AUPRC. A potential explanation for this is that the F1-score targets a specific precision-recall operating point, whereas AUPRC provides a global depiction of the precision-recall tradeoff across all thresholds. Consequently, an IHM may optimize performance at a single decision threshold without improving the overall ranking performance, implying that improvements in F1-score do not necessarily translate to enhanced AUPRC.

AUROC remained consistently high across most IHMs, indicating that overall discrimination power is limited under significant class imbalance regardless of the IHM employed. Reductions in AUROC were primarily restricted to RUS and DMO in specific settings, a finding that aligns with prior investigations.[37,43]

These results highlight that selecting an appropriate IHM must account for both model complexity and target metrics. Specifically, while more powerful models can be improved with IHMs such as ROS and RW, simple models may not enjoy substantial benefits from imbalance handling.

## Limitation

The generalizability of our findings may be constrained by the scope of the public biomedical datasets utilized, which may not fully reflect the diverse and complex nature of broader healthcare contexts or real-world clinical scenarios. Furthermore, our evaluation was limited to a selection of representative IHMs, potentially overlooking specialized or emerging techniques that could offer more nuanced solutions for imbalanced classification. Finally, while we systematically varied data modalities and model architectures, the strategies employed for dataset selection and experimental design may not have captured every intricacy inherent in advanced biomedical data analysis.

## Conclusion

In conclusion, this study offers critical insights into imbalanced biomedical binary classification by systematically examining diverse data modalities, representative imbalance handling methods (IHMs), and models spanning varying levels of complexity. Notably, the effectiveness of IHMs depends on the interplay between model complexity and data modality. While simpler models may not derive significant benefits from imbalance handling, more powerful models can be enhanced by applying appropriate IHMs. Among the evaluated IHMs, random oversampling (ROS) and reweighting (RW) emerged as the most consistently beneficial, particularly for powerful models. Direct F1-score optimization (DMO) demonstrated inconsistent utility, with gains primarily observed for unstructured data. In contrast, random undersampling (RUS) and SMOTE consistently degrade performance and are hence not recommended. These findings underscore that imbalance handling should not be applied uniformly; instead, selection must account for both model complexity and data modality.

# Funding

This work was partially supported by the following funding agencies of the United States: National Cancer Institute (R01CA287413), National Institute of Neurological Disorders and Stroke (R01NS131314), National Science Foundation (IIS 2435911, IIS 2530226), National Center for Complementary and Integrative Health (R01AT009457, U01AT012871), the National Institute on Aging (R01AG078154), the National Institute of Diabetes and Digestive and Kidney Diseases (R01DK115629), Food and Drug Administration (U01FD008720). The authors acknowledge the Minnesota Supercomputing Institute (MSI) at the University of Minnesota for providing resources that contributed to the research results reported within this paper.